# UG-FedDA: Uncertainty-Guided Federated Domain Adaptation for Multi-Center Alzheimer's Disease Detection

*Authors*

Fubao Zhu[a], Zhanyuan Jia[a], Zhiguo Wang[a], Huan Huang[b], Danyang Sun[a], Chuang Han[a], Yanting Li[a], Jiaofen Nan[a], Chen Zhao[b,*], Weihua Zhou[c,d]

*Institutions*

[a]School of Computer Science and Technology, Zhengzhou University of Light Industry, Zhengzhou 450001, Henan, China
[b]Department of Computer Science, Kennesaw State University, Marietta, GA, 30060, USA
[c]Department of Applied Computing, Michigan Technological University, Houghton, MI, 49931, USA
[d]Center for Biocomputing and Digital Health, Institute of Computing and Cybersystems, and Health Research Institute, Michigan Technological University, Houghton, MI, 49931, USA

*Corresponding authors:*

Chen Zhao
Email:czhao4@kennesaw.edu

**Abstract**

Alzheimer's disease (AD) is an irreversible neurodegenerative disorder, and early diagnosis is critical for timely intervention. However, most existing classification frameworks face challenges in multicenter studies, as they often neglect inter-site heterogeneity and lack mechanisms to quantify uncertainty, which limits their robustness and clinical applicability. To address these issues, we proposed Uncertainty-Guided Federated Domain Adaptation (UG-FedDA), a novel multicenter AD classification framework that integrates uncertainty quantification (UQ) with federated domain adaptation to handle cross-site structure magnetic resonance imaging (MRI) heterogeneity under privacy constraints. Our approach extracts multi-template region-of-interest (RoI) features using a self-attention transformer, capturing both regional representations and their interactions. UQ is integrated to guide feature alignment, mitigating source–target distribution shifts by down-weighting uncertain samples. Experiments are conducted on three public datasets: the Alzheimer's Disease Neuroimaging Initiative (ADNI), the Australian Imaging, Biomarkers and Lifestyle study (AIBL), and the Open Access Series of Imaging Studies (OASIS). UG-FedDA achieved consistent cross-domain improvements in accuracy, sensitivity, and area under the ROC curve across three classification tasks: AD vs. normal controls (NC), mild cognitive impairment (MCI) vs. AD, and NC vs. MCI. For NC vs. AD, UG-FedDA achieves accuracies of 90.54%, 89.04%, and 77.78% on ADNI, AIBL and OASIS datasets, respectively. For MCI vs. AD, accuracies are 80.20% (ADNI), 71.91% (AIBL), and 79.73% (OASIS). For NC vs. MCI, results are 76.87% (ADNI), 73.91% (AIBL), and 83.73% (OASIS). These results demonstrate that the proposed framework not only adapts efficiently across multiple sites but also preserves strict privacy. The code is already available on GitHub: https://github.com/chenzhao2023/UG_FADDA_AlzhemiersClassification

# 1. Introduction

Alzheimer's disease is a progressive neurodegenerative disorder characterized by gradual cognitive decline and potentially fatal complications. By 2050, more than 13 million U.S. adults aged 65 years or older are projected to be living with AD, a burden expected to markedly increase healthcare utilization and mortality in older adults [1]. Mild cognitive impairment (MCI) is considered a prodromal stage of AD. The annual conversion rate is approximately 10–15%, and long-term follow-up indicates a cumulative conversion risk exceeding 70% [2]. Given the irreversibility of AD, early MCI screening and timely intervention are primary strategies to curb disease progression, motivating the development of advanced predictive frameworks [3].

Structure magnetic resonance imaging (MRI) is effective in detecting AD-related anatomy abnormalities, including atrophy and other pathological changes, thereby supporting diagnosis [4]. A common MRI workflow employs region-of-interest (RoI) atlases to vectorize images, with regions parcellated by anatomical criteria. The resulting RoI features are modeled with machine learning or deep learning to support AD identification [5]. Single-atlas approaches are limited to a single parcellation and may miss complementary morphological features, whereas integrating multiple atlases can better capture group differences and enhance performance [6]. Early studies often concatenated multi-atlas features while ignoring their interactions. This limitation has motivated the adoption of advanced representation learning techniques that explicitly model dependencies among RoIs across different atlases. In this context, self-attention (SA) [7] has emerged as a powerful mechanism, capable of uncovering complex relationships among RoI features and highlighting the most informative regions for classification tasks. Recent studies have shown that Transformer-based frameworks not only improve multimodal fusion for AD diagnosis [8], but also enhance RoI-level modeling, where attention mechanisms effectively capture both intra- and inter-RoI interactions [9].

Multi-institutional MRI datasets are typically large in scale, offering enhanced model reliability and improved generalization [10]. However, differences in scanner settings, acquisition protocols, and population characteristics introduce inter-site heterogeneity, thereby degrading cross-site generalization [11]. Domain adaptation mitigates this issue by aligning source and target distributions while preserving robustness [12]. Classical approaches typically rely on statistical matching with shallow learners, while deep domain adaptation techniques modify network architectures (e.g., Transformers) or utilize adversarial and discrepancy-based losses to achieve feature alignment. For example, Yao et al. developed a manifold-alignment mechanism to synchronize global and local distributions [13]. Zhu et al. emphasized subdomain matching to improve transfer efficiency [14]. In medical imaging analysis, Yan et al. integrated edge detection into an adversarial framework to enable cross-domain MRI segmentation [15]. Gao et al. used central moment discrepancy to regularize functional MRI classification [16], while Guan et al. designed an attention-guided network to harmonize whole-brain AD diagnosis across different hospitals [17]. However, many of these methods require full data access, raising privacy risks, particularly under the General Data Protection Regulation (GDPR).

Despite recent advances, challenges remain in multi-center AD classification. First, cross-center heterogeneity continues to hinder model generalization; second, direct data sharing poses risks of privacy breaches and regulatory non-compliance. To address these issues, this study proposes an Uncertainty-Guided Federated Domain Adaptation (UG-FedDA) approach for AD classification using cross-centered structure MRI data. First, UQ is employed to evaluate the reliability of samples and features across centers, and UQ-guided feature alignment is applied to reduce distribution discrepancies between source and target domains. Within the Federated Learning (FL) framework, each center conducts local training and shares only model updates. A global aggregation strategy, weighted by uncertainty is then applied to mitigate negative transfer from centers with high uncertainty. Throughout the process, no raw data or labels are exchanged between source and target domains, thereby inherently reducing privacy and compliance risks. Extensive experiments on three multi-center datasets demonstrate that the proposed framework achieves consistent improvements in classification accuracy, sensitivity, and AUC across NC vs. AD, MCI vs. AD, and NC vs. MCI tasks, validating both its effectiveness and clinical potential.

The contributions of this work are as follows:

(1) A federated domain adaptation method that integrates uncertainty with feature alignment by embedding uncertainty estimation into the alignment process. This prioritizes reliable samples and stable features, thereby mitigating source–target distribution discrepancies and enhancing cross-domain discrimination.

(2) An uncertainty-weighted federated learning strategy is developed, in which client contributions are adaptively weighted by uncertainty during global aggregation. This reduces the influence of highly uncertain centers and improves both robustness and generalization.

## 2. Related work

### 2.1 Neuroimaging-based Alzheimer's Disease Classification

Traditional AD classification has primarily relied on machine learning models that use RoI features from structure MRI. Common approaches include support vector machines and random forests, typically based on a single atlas segmentation [18]. However, these approaches are limited because they cannot fully exploit the complementary information provided by multiple atlases [19]. Recent studies have attempted to integrate multi-atlas features, often concatenating them as inputs to deep learning models [20]. To better capture the interactions among RoI features, self-attention mechanisms inspired by Transformers have been increasingly adopted [21]. Originally developed for natural language processing to model long-range dependencies, Transformers have recently shown strong potential in neuroimaging-based Alzheimer's disease classification. For instance, Zhang et al. [22] proposed a Transformer-based multimodal feature fusion framework that integrates MRI and PET data, achieving superior performance over CNN-based fusion approaches on the ADNI dataset. Similarly, Yu et al. [23] designed a Transformer-driven multimodal fusion model and demonstrated consistent improvements in distinguishing AD, mild cognitive impairment (MCI), and normal controls, further highlighting the ability of attention mechanisms to capture cross-modal and inter-regional dependencies. However, cross-domain distribution shifts in multi-site MRI data substantially limit the generalization ability of these models [24].

Domain adaptation (DA) addresses this challenge by aligning source and target domain distributions through techniques such as adversarial loss, local maximum mean discrepancy (LMMD)[16], manifold alignment [25], and subdomain matching [26]. In medical imaging, DA has been widely applied to both MRI segmentation and classification [27]. Nevertheless, existing DA approaches often assume that all samples and features contribute equally, overlooking the variability and noise inherent in multi-center data. This limitation can lead to negative transfers when centers with high distributional discrepancies dominate the alignment process. To address this gap, our work introduces UQ, which explicitly measures the reliability of samples and features, thereby guiding both feature alignment and global aggregation to further enhance cross-domain performance.

### 2.2 Uncertainty Quantification in Medical Imaging

In recent years, UQ has gained significant attention in medical decision support systems and is considered a key approach for improving model reliability [28]. Common methods include Monte Carlo (MC) dropout and Bayesian neural networks, both of which estimate predictive confidence by leveraging randomness [29]. More recently, evidential deep learning has been proposed to directly output Dirichlet distribution parameters, thereby providing class probabilities and predictive uncertainty in a unified framework [30][31]. In medical imaging, UQ has been applied to various tasks, including segmentation [32], lesion detection, and disease classification [33]. The importance of UQ has been further highlighted in recent studies. For example, Zhao et al. introduced HAGMN-UQ for coronary artery semantic labeling, which integrates uncertainty estimation into graph matching to improve structural reliability [34]. Similarly, Mu et al. proposed an uncertainty-aware dynamic decision framework for multi-omics integration, demonstrating that uncertainty can effectively guide feature fusion and decision-making in complex biomedical settings [35]. For AD classification, studies have demonstrated that UQ can highlight ambiguous diagnostic cases (e.g., NC vs. MCI), thereby enhancing interpretability and clinical utility [36]. However, most existing studies treat UQ merely as a post-processing tool during inference, rather than integrating it into training and cross-domain adaptation [37][38]. This limitation underscores the need for developing UQ methods that are not only predictive but also embedded within the training and adaptation process, particularly for AD diagnosis across heterogeneous multi-site data.

## 2.3 Federated Domain Adaptation

FL is a distributed paradigm that supports privacy-preserving multi-center collaboration by training local models at each site and sharing only model updates instead of raw data [39]. In medical imaging, FL is particularly suitable for multi-site MRI analysis because it complies with privacy regulations such as the GDPR [40][41]. FL has been applied to diverse tasks, including multi-site AD classification using fMRI [42], brain segmentation using MRI [43], and retinal disease screening [44].

To address cross-site data heterogeneity (non-IID)—that is, when local data distributions differ significantly across centers due to variations in sample size, modality, or class balance—federated domain adaptation (FedDA) integrates FL with domain adaptation techniques, aligning distributions using approaches such as adversarial training [27], gradient matching [45], and Fourier-style mining [46]. For example, Li et al. [47] proposed a privacy-preserving FedDA framework for multi-site fMRI analysis, while Liu et al. [48] introduced FedDG, which enables federated domain generalization through periodic learning in the frequency domain.

However, FL still faces challenges: non-IID data may cause slow convergence, negative transfer, and instability in minority classes [49]. Recent studies have attempted to alleviate these issues through weighted aggregation [50] or client selection [51]. Nevertheless, contributions from highly uncertain sites can still disrupt the global model [52].In addition, recent advances such as model-level attention and style normalization for federated segmentation in Zhu et al [53]. suggest that incorporating feature stability or uncertainty into the federated learning pipeline may further mitigate cross-site inconsistencies.To address this, we propose uncertainty-weighted federated aggregation, which prioritizes low-uncertainty sites, thereby reducing negative transfer and enhancing cross-domain robustness.

## 3. Data and Methods

### 3.1 Dataset and Preprocessing

Experiments were conducted on three public datasets: (1) the Alzheimer's Disease Neuroimaging Initiative (ADNI); (2) the Australian Imaging, Biomarkers and Lifestyle study (AIBL); and (3) the Open

Access Series of Imaging Studies (OASIS). Baseline T1-weighted structural MRI scans were used for all subsequent analyses.

Samples from the three datasets were categorized into three diagnostic groups: AD, MCI and NC. 395 participants from ADNI dataset were enrolled (AD = 102, MCI = 100, NC = 193). AIBL included 395 participants (AD = 74, MCI = 104, NC = 217). OASIS included 556 participants (AD = 225, MCI = 71, NC = 260). Table 1 describes the demographic information for the enrolled subjects.

Table 1. Demographic information of the enrolled subjects across three datasets

| Dataset | Category | Age | Sex(M/F) | MMSE | CDR-SB |
|---|---|---|---|---|---|
| ADNI | AD | 74.27 ± 7.78 | 58/44 | 22.97 ± 2.08 | 4.56 ± 1.61 |
| | MCI | 74.24 ± 7.74 | 68/32 | 27.27 ± 1.66 | 1.44 ± 0.70 |
| | NC | 74.21 ± 5.33 | 102/91 | 29.03 ± 1.13 | 0.03 ± 0.14 |
| AIBL | AD | 74.39 ± 8.01 | 28/46 | 20.47 ± 5.64 | - |
| | MCI | 74.96 ± 7.20 | 53/51 | 27.05 ± 2.17 | - |
| | NC | 73.41 ± 7.16 | 88/128 | 28.76 ± 1.26 | - |
| OASIS | AD | 74.68 ± 7.46 | 127/98 | 24.81 ± 3.95 | 3.52 ± 2.03 |
| | MCI | 74.74 ± 6.09 | 32/39 | 27.61 ± 2.15 | 1.24 ± 0.76 |
| | NC | 67.85 ± 12.94 | 104/156 | 29.25 ± 0.94 | 0.02 ± 0.09 |

The preprocessing pipeline for structural MRI was as follows: First, SPM12 [54] is used to perform anterior commissure–posterior commissure (AC–PC) alignment to ensure consistent head orientation and register images to a standard coordinate system, thereby reducing inter-subject angular variance that could bias classification [55]. Next, CAT12 [56] is applied for structured preprocessing, including skull stripping to remove scalp and bone confounds, followed by bias-field correction to mitigate intensity inhomogeneity and homogenize tissue contrast [57]. For tissue segmentation, CAT12 further partitioned the images into gray matter (GM), white matter (WM), and cerebrospinal fluid (CSF). Because AD primarily affects GM volumetry, only the segmented GM images are retained to capture disease-relevant features more precisely. The GM images are then spatially normalized to the Montreal Neurological Institute (MNI) [58] reference space, harmonizing inter-subject anatomical differences and facilitating stable feature learning across centers. To compensate for voxel-wise changes introduced by normalization, modulation is applied to preserve volumetric information across subjects and stabilize the data distribution.

The Data Processing Assistant for Resting-State fMRI (DPARSF) toolbox [59] is used to extract RoI features from the preprocessed, normalized gray-matter (GM) images. The Brodmann brain atlas used in this study is obtained from the DPARSF toolbox. For each atlas, we computed a feature vector comprising the mean GM volume per region. Consequently, each subject contains eight vectors (one per atlas and the detailed information is shown in Table 2).

Table 2. Information of different employed templates

| Template | Number of regions |
|---|---|
| AAL1 [60] | 116 |
| AAL2 [60] | 120 |
| AAL3 [60] | 166 |
| Brainnetome [61] | 246 |
| Brodmann | 41 |
| Hammersmith [62] | 83 |
| HarvardOxford [63] | 67 |

| Juelich [63] | 121 |

Because vector lengths vary across atlases, all vectors are zero-padded to the maximum length (246). After extracting the RoI features from the eight atlas, eight RoI feature vectors (one per atlas) are organized as a sequence, yielding a unified 8 × 246 input matrix.

### 3.2 Methods

We propose a multi-center Alzheimer's disease classification method that integrates uncertainty quantification with federated domain adaptation. The overall framework is illustrated in Figure 1. Three public datasets—ADNI, AIBL, and OASIS—serve as Centers A, B, and C, respectively. Each dataset is alternately used as the target domain, with the remaining two as sources, forming a two-source, one-target configuration that enhances domain-invariant representation learning and generalization to unseen centers.

In the Uncertainty-Aware Feature Style Alignment Module, each center first employs the shared Style Feature Extractor to obtain style features from its local data. To assess feature stability, we perform multiple forward passes with Monte Carlo (MC) Dropout, compute the mean and variance of features, and use the variance as an uncertainty score. This variance reflects the model's confidence: higher variance indicates greater uncertainty due to inconsistent feature representations across passes.

Under federated learning (FL), raw data remain local while only model parameters or feature statistics are shared, ensuring privacy. In the Uncertainty-Aware Feature Style Alignment Module, each center extracts local style features using a shared feature extractor. Monte Carlo dropout generates feature means and variances, with the latter representing uncertainty. These uncertainty scores weight reliable features during alignment. Centers upload the mean and uncertainty of their features, and the server aligns source and target domains via an uncertainty-weighted KL divergence loss to reduce cross-center shifts.

Two alignment extractors (AC and BC) learn mappings between sources (A, B) and the target (C), with C serving as a bridge to indirectly align A and B. The extractors are aggregated using uncertainty weighting to form a robust global model. Finally, in the Uncertainty-Aware Classification Module, local representations are concatenated with aligned style features and trained using evidential learning. Uncertainty-weighted federated aggregation combines local models, enabling robust and privacy-preserving multi-center knowledge integration.

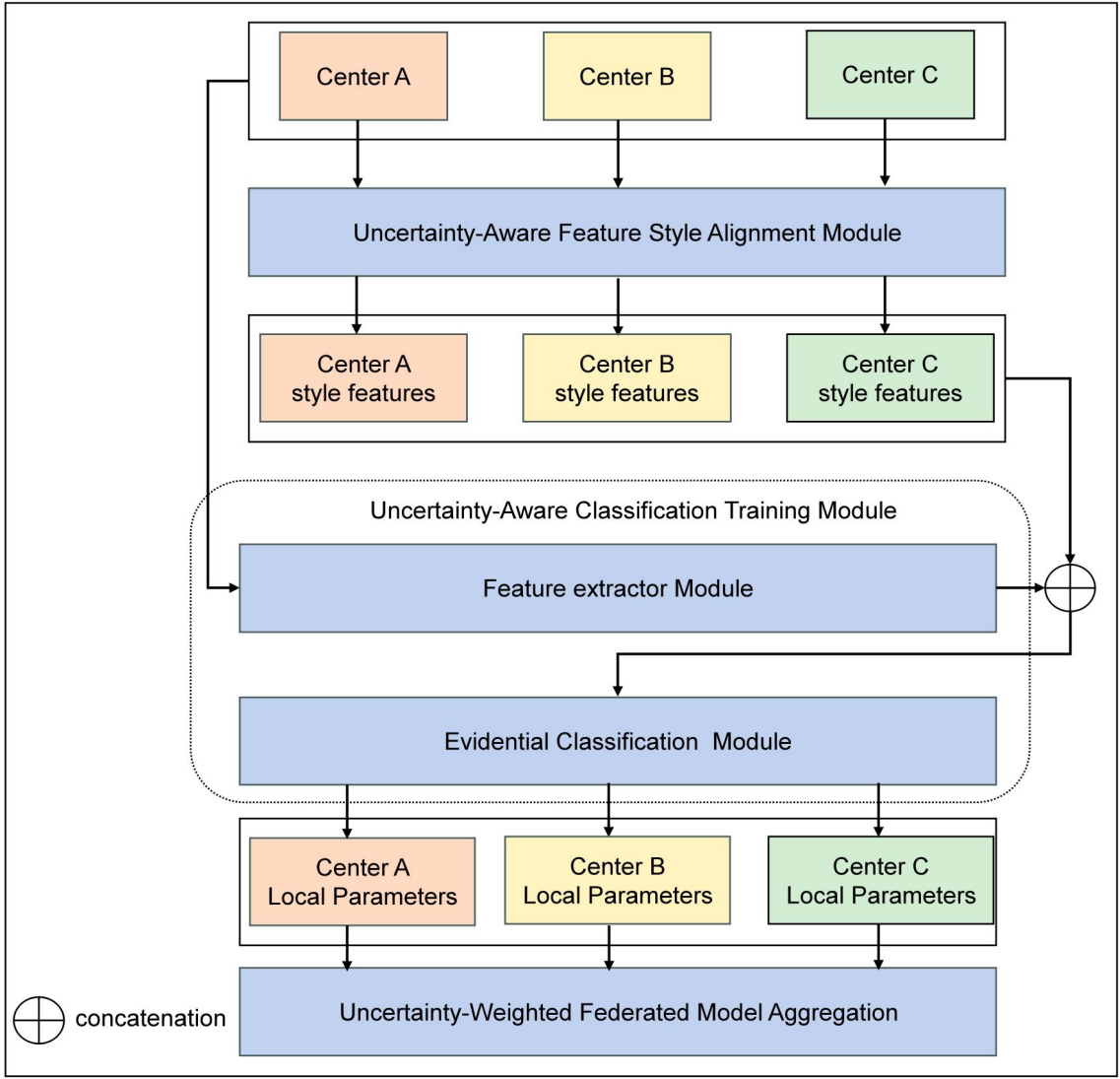

Figure 1. Framework of the proposed Alzheimer's disease diagnosis method based on UQ and federated domain adaptation

### 3.2.1 Uncertainty-Aware Feature Style Alignment Module

In cross-domain AD classification, distributional discrepancies between source and target domains primarily arise from the heterogeneity of multi-site MRI data, including differences in scanners, acquisition protocols, and subject demographics. This heterogeneity remains reflected in the feature distributions even after ROI extraction, leading to significant statistical discrepancies in ROI features across different centers. Such domain shifts substantially impair model generalization, rendering models trained on one site unreliable when applied to others. Traditional feature alignment methods, such as statistical matching [64] or adversarial training [65], can mitigate part of these discrepancies. However, they typically ignore uncertainty in the data, such as low-confidence samples caused by noise, blurred anatomical boundaries, or ambiguous labels. Indiscriminate alignment of such samples may lead to negative transfer or overfitting. To address this limitation, we introduce UQ as a guiding mechanism, enabling the model to dynamically distinguish reliable from unreliable samples during cross-domain alignment and thereby enhancing the robustness of cross-site learning.

To this end, we designed a feature style extractor that integrates Transformers with convolutional neural networks (CNNs), as illustrated in Figure 2 (a). The module takes as input a sequence of RoI features extracted from multiple templates, with shape (B, 8, 246), where B denotes the batch size, 8 is the number of brain templates, and 246 is the standardized feature length after zero-padding.

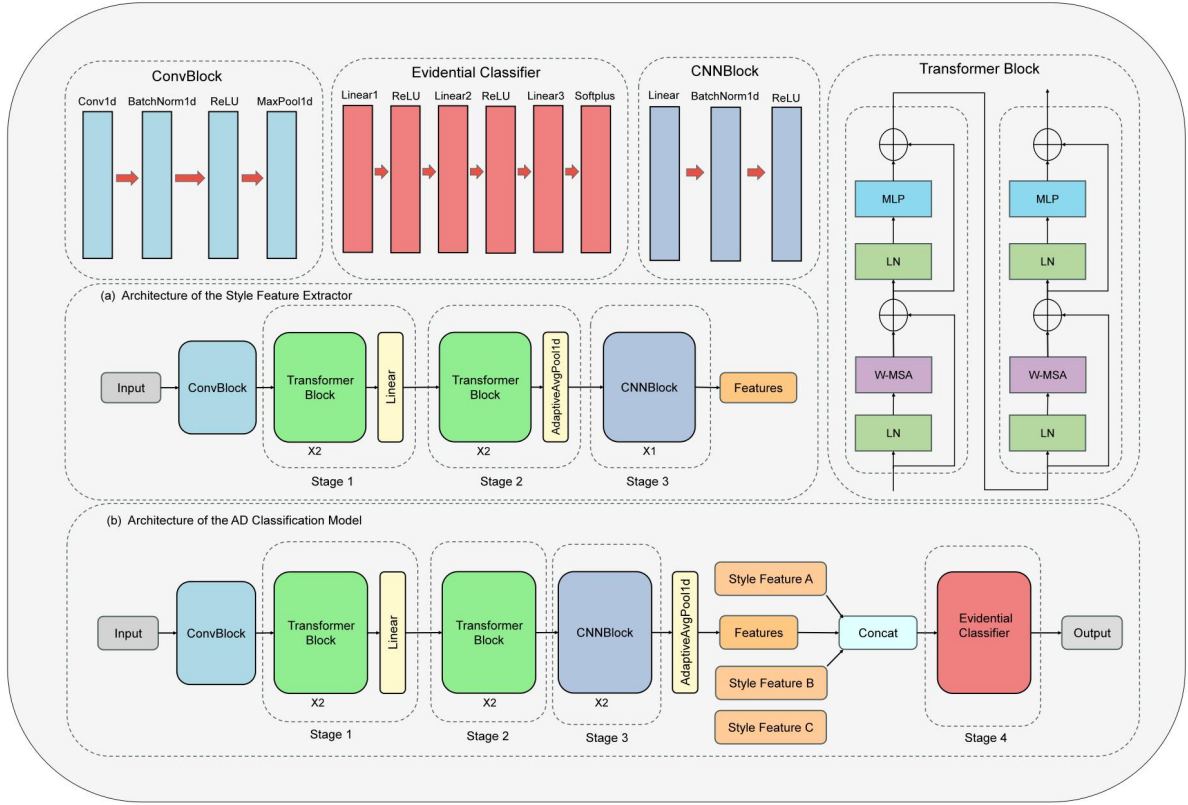

Figure 2. Model Architecture

The input sequence $x \in R^{B\times 4\times 128}$ first passes through a one-dimensional convolution, batch normalization, and ReLU activation, followed by max pooling that reduces the sequence length from 8 to 4:

$$H^{(0)} = MaxPool(ReLU(BN(Conv1D(x)))), \quad H^{(0)} \in R^{B\times 4\times 128} \quad (1)$$

To capture both local and global features, as well as the relational structures among brain regions, the resulting features are processed by two Transformer blocks, which are employed to perform feature embedding, as shown in Eq. 2.:

$$H^{(1)} = TransformerBlock(H^{(0)}), \quad H^{(1)} \in R^{B\times 4\times 128} \quad (2)$$

A patch merging operation is subsequently performed to concatenate neighboring feature embeddings and project them into a higher-dimensional space, as shown in Eq. 3.

$$H^{(2)} = PatchMerge(H^{(1)}) \in R^{B\times 2\times 256} \quad (3)$$

Subsequently, two additional Transformer Blocks with the dimension of 256 are applied to $H^{(2)}$, capturing long-range dependencies and refining semantic representations, as shown in Eq. 4.

$$H^{(3)} = TransformerBlock(H^{(2)}) \in R^{B\times 2\times 256} \quad (4)$$

Finally, adaptive average pooling is applied along the sequence length to compress the representations into a single vector, which aggregates information across all RoIs and produces a fixed-length feature representation suitable for downstream analysis, as shown in Eq. 5.

$$H^{(4)} = AdaptiveAvgPool(H^{(3)}) \in R^{B\times 256} \quad (5)$$

The extracted feature $H^{(4)}$ is subsequently passed through a CNN block composed of a linear layer, batch normalization, and ReLU activation to generate the final compact style representation, as shown in Eq. 6.

$$z = ReLU(BN(Linear(H^{(4)}))), \quad z \in R^{B \times 256} \tag{6}$$

During the feature extraction and alignment process, we employ the MC Dropout sampling mechanism to estimate feature-level uncertainty. For each input sample $x$, multiple stochastic forward passes are performed under the MC Dropout framework, where a subset of neurons is randomly deactivated in each pass, effectively sampling different sub-networks from the full model. After multiple forward passes, the feature representations of the same sample are averaged to obtain a stable feature representation $z_i$, where $i$ denotes the sample index. To obtain a domain-level style representation, we computed the average style features across samples in each domain and their within-domain variance, yielding a domain-level style feature and an associated domain-level uncertainty measure, denoted by $\mu_d$ and $\sigma_d^2$ in Eq. 7.

$$\mu_d = \frac{1}{N_d}\sum_{i=1}^{N_d} z_i^{(d)}, \quad \sigma_d^2 = \frac{1}{N_d}\sum_{i=1}^{N_d}\left(z_i^{(d)} - \mu_d\right)^2, \quad d \in \{s, t\} \tag{7}$$

where, $d \in \{s, t\}$ represents the source domain ($s$) and the target domain ($t$), respectively. $N_d$ denotes the number of samples in domain d, $\mu_d$ denotes the domain-level style feature vector, and $\sigma_d^2$ denotes the domain-level uncertainty measure.

During domain alignment, a Kullback–Leibler (KL) divergence–based loss was employed to reduce the style-distribution gap between the source and target domains. The KL divergence is a well-established statistical measure that quantifies the dissimilarity between two probability distributions. This equation is derived from the standard definition of KL divergence, as shown in Eq. 8.

$$D_{KL}(p_t \| p_s) = \int p_t(x) \log \frac{p_t(x)}{p_s(x)} dx \tag{8}$$

where $p_t(x)$ denotes the distribution of the sample x in target domain.

Since the domain-level style features are obtained through feature aggregation across numerous samples, their overall distribution can be reasonably approximated by a Gaussian according to the central limit theorem. The aggregated representations from each domain tend to follow a normal distribution characterized by a domain-specific mean and variance. We model the source and target style feature prior distributions as Gaussian densities $p_s \sim N(\mu_s, \sigma_s^2)$ and $p_t \sim N(\mu_t, \sigma_t^2)$, where the probability density function is defined in Eq. 9.

$$p(x) = \frac{1}{\sqrt{2\pi\sigma^2}} e^{-\frac{(x-\mu)^2}{2\sigma^2}} \tag{9}$$

By substituting these Gaussian densities into the KL divergence formulation in Eq. 9 and carrying out the integration, we obtain the analytical form as shown in Eq. 10

$$L_{KL} = \frac{1}{2}\left[\frac{\sigma_s^2}{\sigma_t^2}(\mu_t - \mu_s)^2 + \log\left(\frac{\sigma_t^2}{\sigma_s^2}\right)\right] \tag{10}$$

This formulation directly originates from the closed-form KL divergence between two Gaussian distributions and enables simultaneous alignment of mean and variance across domain-level style representations, promoting consistent and stable adaptation across centers.

### 3.2.2 Uncertainty-Aware based AD Classification

In the AD classification task, traditional methods often rely solely on single-sample features while neglecting the impact of cross-domain distributional discrepancies on model generalization. To enhance the robustness and clinical applicability of the model, we design a classifier that integrates sample features with domain-level style features. However, data collected from different centers typically exhibit noise and ambiguity. For instance, the boundary between MCI and AD is often indistinct, causing Softmax-based classifiers to generate overconfident predictions that may lead to potential misdiagnoses. To obtain calibrated confidence representations at the prediction level, we employ a Dirichlet-based uncertainty modeling approach. Specifically, the network outputs are regarded as evidence, and a Dirichlet distribution is used to model the probability distribution over classes. Let the network output be the evidence vector $e = [e_1, ..., e_K]$, where the corresponding Dirichlet parameters are defined as $\alpha_k = e_k + 1$, The Dirichlet strength is given by $S = \sum_{k=1}^{K} \alpha_k$, the confidence for class k is $p_k = \frac{\alpha_k}{S}$ and the uncertainty is $u = \frac{K}{S}$, where $K$ denotes the total number of classes and $k$ is the class index. In our implementation, the Softplus activation function is employed to ensure non-negative outputs, which are interpreted as evidence for the Dirichlet distribution during prediction.

Meanwhile, to reduce the randomness of single forward passes and improve the stability of predictions, MC Dropout is introduced into the model. During inference, dropout remains active, and multiple stochastic forward passes are performed, with their results averaged to obtain the final feature representation. This multi-sampling mechanism effectively mitigates the instability caused by random initialization or noise perturbations, enabling the model to maintain consistent decision outcomes across multiple inferences and thereby further enhancing the overall prediction stability.

As illustrated in Fig. 2(b), the overall model architecture first extracts sample features $f$ from the input data through multiple stages of feature encoding. Meanwhile, style features $feat_a$, $feat_b$, $feat_c$ are extracted from the data of three centers, respectively. These style features represent the overall distributional patterns of different centers in the feature space and reflect the domain-specific characteristics of each center. By integrating sample features with multiple domain-level style features, the model can adaptively align feature representations across domains, thereby enhancing domain adaptability and classification robustness. Subsequently, the sample feature $f$ is concatenated with the style features $feat_a$, $feat_b$, $feat_c$ to form an extended input: $f_{cat} = [f, feat_a, feat_b, feat_c]$, with a dimension of $1 \times 1024$, The extended feature input $f_{cat}$ is fed into the Evidential Classifier, which passes through three fully connected layers and uses the Softplus activation to produce the evidence vector $e$. During inference, the evidence is converted into Dirichlet distribution parameters $\alpha = [\alpha_1, ..., \alpha_K]$, and the class-wise confidence $p_k = \frac{\alpha_k}{S}$ and uncertainty are computed $u = \frac{K}{S}$, where $S = \sum_{k=1}^{K} \alpha_k$. Finally, the class with the highest confidence $p_k$ is selected as the model prediction result.

During classification training, the loss function is composed of the Uncertainty Cross Entropy Loss ($L_{UCE}$) and the Mean Squared Error (MSE) consistency loss ($L_{MSE}$). The total loss is defined in Eq. 11.

$$L_{loss} = \frac{1}{2}(L_{UCE} + L_{MSE}) \tag{11}$$

where, $L_{UCE}$ is based on the Dirichlet distribution, as shown in Eq. 12.

$$L_{UCE} = \sum_{k=1}^{K} y_k \left(\psi(S) - \psi(\alpha_k)\right) \tag{12}$$

where, $\psi(.)$ denotes the digamma function, $\alpha_k$ is the Dirichlet parameter associated with class k, $S$ is Dirichlet strength and $y_k$ is the one-hot label. The loss is uncertainty-aware: it emphasizes reliable samples, suppresses the influence of noisy or uncertain ones, and indirectly stabilizes the decision boundary around confident samples, thereby enhancing cross-domain robustness.

Additionally, we introduce $L_{MSE}$ to maintain feature stability during the model optimization, as shown in Eq. 13.

$$L_{\text{MSE}} = \|\mu_{\text{prev}} - \mu_{\text{curr}}\|^2 \tag{13}$$

where, $\mu_{\text{curr}}$ denotes the mean feature of the current round, and $\mu_{\text{prev}}$ denotes the mean feature from the previous round; for the first round, $\mu_{\text{prev}}$ is set to 0. This term enforces the consistency of the mean feature representations across communication rounds, effectively alleviating potential distributional shifts that may arise during local updates. By stabilizing the evolution of feature statistics, it smooths the training dynamics and facilitates stable and efficient convergence in the federated optimization process.

### 3.2.3 Uncertainty-Weighted Federated Model Aggregation

In the FL framework, multi-site MRI data often display non-independent and identically distributed (non-IID) characteristics, indicating that sample distributions differ substantially across centers. For instance, some centers may contribute lower-quality or noisier samples, while others may be biased toward particular populations. These discrepancies cause high-uncertainty centers to introduce "negative transfer" during model aggregation, thereby slowing global model convergence, reducing stability, and weakening cross-domain generalization. Traditional aggregation methods (e.g., FedAvg) assign weights to centers solely based on sample size, disregarding data quality and uncertainty. This limitation is especially critical in medical imaging tasks, where ambiguous diagnostic boundaries or noisy labels further amplify inter-domain discrepancies.

To address this issue, we propose an uncertainty-weighted aggregation strategy. In the federated learning framework, after each training round, individual centers upload their locally trained model parameters, which are aggregated into a global model using uncertainty-based weighting. Because uncertainty scores are computed for all centers in every round, this information can be directly used to adjust model weights, thereby improving the stability of the global model and enhancing its cross-domain adaptability.

Let $\theta_t$ denote the sample set of the $t$-th data center. Before aggregation, we summarize the uncertainty scores from all centers and compute the mean uncertainty, as shown in Eq. 14.

$$u_c = \frac{1}{|X_c|} \sum_{i \in X_c} u_i \tag{14}$$

Here, $X_c$ denotes the sample set at data center c; $u_i$ denotes the uncertainty score of samples.

To prioritize low-uncertainty sites and down-weight high-uncertainty sites, we employed an uncertainty-weighted aggregation scheme that dynamically scaled each site's model updates when forming the global model. First, we computed uncertainty-based weights (coefficients) as shown in Eq. 15.

$$w_t = w_t^{FedAvg} \cdot \frac{1}{1 + \exp\left(-\gamma(u_t - \tau)\right)} \tag{15}$$

Here, $w_t^{FedAvg} = \frac{|X_t|}{\Sigma_j |X_j|}$ denotes the standard federated-learning weighting coefficient; $\gamma$ controls the steepness of the weight decay; and $\tau$ is the site-level uncertainty threshold used to dynamically adjust aggregation.

Finally, the model parameters are updated as in Eq. 16.

$$\theta^{global} = \sum_t w_t \theta_t \quad (16)$$

The updated model parameters $\theta^{global}$ were broadcast to all sites and used to initialize the next training round. In our classification tasks, the model outputs a probability distribution over two diagnostic categories, corresponding to one of three binary settings: AD vs NC, MCI vs AD, or NC vs MCI. The ground truth labels are directly taken from the clinical diagnoses provided in each dataset. During aggregation, the uncertainty-weighted scheme prioritizes contributions from low-uncertainty samples, reduces the influence of high-uncertainty samples, stabilizes the global model, and thereby improves the robustness of federated learning in cross-domain scenarios.

## 4. Experiments

### 4.1 Experimental Setup

Experiments were conducted using PyTorch on an NVIDIA Tesla V100 GPU. We used the Adam optimizer with an initial learning rate of $1\times10^{-4}$ and a batch size of 32. The style-alignment stage was trained for 25 epochs, and the classification stage for 100 epochs. The number of MC Dropout forward passes is set to 10 for uncertainty estimation.

### 4.2 Evaluation Metrics

In medical image classification, evaluation metrics quantify model performance across tasks. Because these models support clinical applications—such as disease prediction and lesion detection—stringent accuracy and stability are required. Common metrics include precision, sensitivity/recall, accuracy (ACC), and the area under the receiver operating characteristic curve (AUC).

## 5. Results

### 5.1 Domain Adaptation Performance Evaluation

We benchmarked against FACT [66], FedADA [67], and DBE [68] on three cross-site binary tasks: NC vs. AD, MCI vs. AD, and NC vs. MCI. Overall, our method achieved the most consistently high accuracy (ACC) in 8 of the 9 comparisons.

(1) NC vs. AD

Across the three evaluation sites ADNI, AIBL and OASIS, our method achieved accuracies of 90.54%, 86.85%, and 78.76%, respectively, the highest among all compared methods. Although our method did not produce the highest AUC on every site, the differences were generally small. On AIBL, FedADA obtained the highest AUC at 88.66%, while our method achieved 82.17%. On ADNI, FACT reached 90.94%, FedADA achieved 91.22%, and DBE reached 91.61%, whereas our method obtained the highest value at 91.68%. On OASIS, our method delivered a sensitivity of 78.03%, substantially higher than the competing methods, including FACT at 50.18%, FedADA at 45.56%, and DBE at 62.32%. These results highlight the strong sensitivity advantage of UG-FedDA on the most challenging site and demonstrate the model's ability to provide consistently strong performance across domains.

Table 3. Performance Comparison of Different Methods on the NC vs. AD Task (five repeated experiments)

| Task | Method | ACC | P | SEN | P | AUC | P |
|---|---|---|---|---|---|---|---|
| ADNI | FACT | 86.62±0.88 | <0.001 | 73.06±1.71 | 0.018 | 90.94±0.23 | 0.001 |
| | FedADA | 87.84±0 | <0.001 | 73.47±0 | 0.025 | 91.22±0 | 0.015 |
| | DBE | 86.08±1.13 | <0.001 | 78.59±3.35 | 0.654 | 91.61±0.14 | 0.586 |

| | UG-FedDA | **90.95±0.6** | - | **80.82±4.70** | - | **91.68±0.25** | - |
|---|---|---|---|---|---|---|---|
| AIBL | FACT | 85.62±0 | 0.088 | 63.41±0 | 0.374 | 86.05±0.17 | <0.001 |
| | FedADA | **87.67±0** | 0.210 | **82.93±0** | 0.001 | **88.66±0** | <0.001 |
| | DBE | 81.78±0.61 | <0.001 | 64.87±2.18 | 0.216 | 84.32±0.14 | 0.001 |
| | UG-FedDA | 86.85±1.22 | - | 60.49±6.54 | - | 82.17±0.63 | - |
| OASIS | FACT | 72.92±0.79 | <0.001 | 50.18±1.72 | <0.001 | **83.80±1.52** | 0.605 |
| | FedADA | 74.90±0 | <0.001 | 45.56±0 | <0.001 | 81.72±0 | 0.030 |
| | DBE | 74.65±0.62 | <0.001 | 62.32±2.71 | <0.001 | 82.02±0.54 | 0.053 |
| | UG-FedDA | **78.76±0.62** | - | **78.03±2.79** | - | 83.35±1.10 | - |

(2) NC vs. MCI

We achieved the highest accuracy ACC on ADNI, AIBL and OASIS, reaching 77.69%, 72.75%, and 83.49% respectively. On ADNI, the margin over the strongest baseline remained meaningful, since our ACC reached 77.69% while the best competing method achieved 74.15%. For sensitivity, our method delivered the best results on AIBL and ADNI, obtaining 40.33% and 50.00%, which demonstrates a stronger ability to detect early impairment patterns. Although our AUC was not always the highest, such as on AIBL where we obtained 69.63% compared with 69.50% from FACT, the results indicate that threshold independent ranking still has room for improvement. This behavior reflects the subtle imaging signatures of MCI and strong cross site variation, where the model tends to balance recall against stability in ranking metrics. For the finer NC versus MCI distinction, further improvement is needed to better reconcile discriminative ability with probabilistic calibration.

Table 4. Performance Comparison of Different Methods on the NC vs. MCI Task (five repeated experiments)

| Task | Method | ACC | P | SEN | P | AUC | P |
|---|---|---|---|---|---|---|---|
| ADNI | FACT | 73.20±0.61 | <0.001 | 37.50±2.08 | 0.009 | **76.04±0.20** | 0.627 |
| | FedADA | 74.15±0 | <0.001 | 33.33±0 | 0.004 | 69.40±0 | 0.001 |
| | DBE | 73.47±1.27 | <0.001 | 30.42±5.63 | <0.001 | 70.75±0.87 | 0.001 |
| | UG-FedDA | **77.69±0.57** | - | **50.00±6.25** | - | 75.64±1.72 | - |
| AIBL | FACT | 72.05±0.62 | 0.136 | 30.84±1.57 | 0.072 | 69.50±0.38 | 0.932 |
| | FedADA | 70.19±0 | 0.001 | 21.05±0 | 0.008 | 65.71±0 | 0.044 |
| | DBE | 69.07±0.28 | <0.001 | 14.74±0.96 | 0.002 | 64.14±1.20 | 0.012 |
| | UG-FedDA | **72.75±0.71** | - | **40.33±8.74** | - | **69.63±3.02** | - |
| OASIS | FACT | 80.24±0.79 | <0.001 | 30.59±1.61 | 0.197 | **72.49±0.33** | 0.089 |
| | FedADA | 81.93±0 | 0.003 | **41.18±0** | 0.005 | 69.59±0 | 0.999 |
| | DBE | 80.12±2.45 | 0.035 | 28.82±5.26 | 0.524 | 67.49±0.71 | 0.183 |
| | UG-FedDA | **83.49±0.54** | - | 26.47±5.88 | - | 68.59±2.90 | - |

(3) MCI vs. AD

Across all three sites, UG-FedDA consistently delivered the strongest performance on the MCI vs AD task. On ADNI, the model achieved 79.21% accuracy, 80.00% sensitivity, and 82.82% AUC, outperforming all competing methods by a clear margin. On AIBL, UG-FedDA also reached the highest accuracy at 72.13% and achieved 72.00% AUC, demonstrating effective discrimination under substantial cross-site variability. The most striking improvement appeared on OASIS, where the model achieved 79.46% accuracy and an exceptionally high sensitivity of 99.32%, reflecting its strong capability to detect mild impairment even in challenging domains. Although the AUC on OASIS was lower at 61.01%, the combination of high accuracy and nearly perfect sensitivity suggests that UG-FedDA enhances progression-related signal separation under severe domain shift. Overall, the results indicate that integrating style alignment with uncertainty modeling improves both threshold-based metrics such as accuracy and sensitivity and threshold-free ranking through AUC, reinforcing its value for robust cross-domain progression identification.

Table 5. Performance Comparison of Different Methods on the MCI vs. AD Task (five repeated experiments)

| Task | Method | ACC | P | SEN | P | AUC | P |
|---|---|---|---|---|---|---|---|
| ADNI | FACT | 69.31±1.40 | <0.001 | 60.37±2.81 | 0.001 | 80.05±0.23 | 0.005 |
| | FedADA | 67.33±0 | <0.001 | 62.96±0 | 0.005 | 67.93±0 | <0.001 |
| | DBE | 66.54±2.26 | <0.001 | **84.44±6.49** | 0.314 | 78.80±0.97 | <0.001 |
| | UG-FedDA | **79.21±1.57** | - | 80.00±6.60 | - | **82.82±1.12** | - |
| AIBL | FACT | 62.47±1.28 | <0.001 | 70.59±3.60 | 0.608 | 65.94±0.91 | <0.001 |
| | FedADA | 66.29±0 | 0.003 | 82.35±0 | 0.001 | 70.32±0 | 0.043 |
| | DBE | 53.48±6.29 | 0.002 | **89.41±5.74** | <0.001 | 65.65±0.50 | <0.001 |
| | UG-FedDA | **72.13±2.01** | - | 69.41±3.35 | - | **72.00±1.29** | - |
| OASIS | FACT | 49.59±1.13 | <0.001 | 43.56±1.42 | <0.001 | 60.96±0.16 | 0.893 |
| | FedADA | 69.59 | <0.001 | 81.36 | <0.001 | 58.81 | 0.500 |
| | DBE | 55±5.16 | <0.001 | 52.55±9.47 | <0.001 | 59.99±0.71 | 0.812 |
| | UG-FedDA | **79.46±0.57** | - | **99.32±1.10** | - | **61.01±5.44** | - |

For AUC and sensitivity SEN, improvements were most evident on the MCI vs AD task. AUC on the NC vs MCI task was lower, showing remaining space for improvement. Most method differences were statistically significant p < 0.05, while some comparisons, such as NC vs AD AUC on ADNI, were not significant.

## 5.2 Ablation study

**Controbution of different modules:**

The proposed UG-FedDA framework is built upon three complementary components designed to enhance cross-domain robustness for multi-site Alzheimer's disease classification. First, the feature ftyle alignment module (F) reduces inter-site distribution shifts by aligning latent feature distributions, thereby improving cross-domain consistency. Second, the Transformer module (T) captures long-range dependencies across ROI feature dimensions and models inter-region relationships that conventional CNN or MLP layers cannot represent, providing richer contextual representation. Third, the evidential classifier (E) introduces uncertainty-aware prediction, allowing the model to estimate evidence and uncertainty jointly, which leads to more calibrated and reliable outputs.

As shown in Table 6, a comprehensive ablation study was conducted on the ADNI dataset to quantify the contribution of each module to overall model performance. The baseline model exhibited limited discriminative capability, achieving 71.29% accuracy, 77.78% sensitivity, and 76.95% AUC on the MCI vs AD task. Performance on NC vs AD and NC vs MCI was similarly constrained, with AUC values of 89.89% and 67.97% and sensitivities of 48.98% and 43.75%, respectively, indicating insufficient robustness under cross-domain variability. Incorporating the feature alignment module baseline+F led to consistent gains across all tasks, increasing the MCI vs AD AUC to 80.91%, and improving NC vs AD sensitivity and AUC to 77.55% and 91.29%. The NC vs MCI AUC also rose to 72.93%, demonstrating enhanced cross-site generalization. Introducing the Transformer module baseline+T further strengthened representational capacity and dependency modeling. This configuration achieved 73.27% accuracy for MCI vs AD and markedly improved NC vs AD sensitivity to 85.71%, accompanied by an AUC of 90.97%. NC vs MCI performance also increased modestly to 72.79% accuracy. The evidential classifier baseline+E improved uncertainty modeling and predictive calibration. It yielded an MCI vs AD AUC of 80.61% and produced the highest NC vs AD sensitivity among all single-module settings at 89.80%. It also achieved the best NC vs MCI AUC among single-module variants at 76.54%, reflecting improved decision reliability under distribution shift. Combining feature alignment and evidential learning baseline+F+E provided a balanced enhancement in discriminability and uncertainty calibration. The MCI vs AD AUC reached 86.05%, while NC vs AD achieved 91.75% AUC. For NC vs MCI, accuracy and AUC increased to 74.83% and 69.00%. The joint use of feature alignment and the Transformer module baseline+F+T further amplified performance, producing 77.23% accuracy for MCI vs AD and the highest NC vs AD AUC across all two-module configurations at 92.76%. This combination also improved NC vs MCI accuracy to 76.19%, demonstrating complementary benefits between alignment and attention-based representation. The full UG-FedDA model, integrating all three components, achieved the most consistent and superior performance across all tasks. It attained 80.20% accuracy, 83.33% sensitivity, and 83.57% AUC for MCI vs AD; 90.54% accuracy, 83.67% sensitivity, and 91.34% AUC for NC vs AD; and 77.55% accuracy, 56.25% sensitivity, and 73.46% AUC for NC vs MCI. These results demonstrate that jointly optimizing feature alignment, Transformer-based representation, and evidential uncertainty learning yields the most robust generalization and the strongest capability for modeling subtle disease progression patterns.

Table 6. Ablation experiments between modules

| Dataset | Model | MCI VS AD | | | NC vs. AD | | | NC vs MCI | | |
|---|---|---|---|---|---|---|---|---|---|---|
| | | ACC | SEN | AUC | ACC | SEN | AUC | ACC | SEN | AUC |
| ADNI | baseline | 71.29 | 77.78 | 76.95 | 82.43 | 48.98 | 89.89 | 70.07 | 43.75 | 67.97 |
| | baseline+F | 79.21 | 81.48 | 80.91 | 88.51 | 77.55 | 91.29 | 72.11 | 20.83 | 72.93 |
| | baseline+T | 73.27 | 70.37 | 76.52 | 86.49 | 85.71 | 90.97 | 72.79 | 22.92 | 66.88 |
| | baseline+E | 75.25 | 72.22 | 80.61 | 83.11 | **89.80** | 90.50 | 74.15 | 33.33 | **76.54** |
| | baseline+F+E | 76.24 | 62.96 | **86.05** | 89.86 | 71.43 | 91.75 | 74.83 | 27.08 | 69.00 |
| | baseline+F+T | 77.23 | 79.63 | 82.23 | 89.19 | 79.59 | **92.76** | 76.19 | 33.33 | 75.42 |
| | UG-FedDA | **80.20** | **83.33** | 83.57 | **90.54** | 83.67 | 91.34 | **77.55** | **56.25** | 73.46 |

**Contribution of different loss functions:**

Table 7 summarizes the ablation results for different loss function configurations on the ADNI dataset. Models trained with only $L_{KL}$ or $L_{MSE}$ exhibited limited discriminability, reflected by MCI vs AD AUC below 62% and consistently low performance on NC vs MCI. Incorporating the uncertainty-aware loss $L_{UCE}$ substantially enhanced separability and calibration, yielding an MCI vs AD AUC of 82.43% and an NC vs AD AUC of 90.87%. Pairwise combinations provided complementary benefits: $L_{UCE}+L_{MSE}$ achieved the highest NC vs AD AUC among two-loss configurations at 91.75%, while $L_{KL}+L_{UCE}$ delivered the strongest

NC vs MCI AUC at 75.74%, indicating that KL regularization becomes effective primarily when guided by uncertainty-aware learning. The full integration of the three losses $L_{KL} + L_{UCE} + L_{MSE}$ yielded the most consistent and robust performance across all tasks, reaching 80.20% accuracy and 83.57% AUC on MCI vs AD, together with 77.55% accuracy and 56.25% sensitivity on NC vs MCI. These results demonstrate that jointly optimizing feature alignment, uncertainty modeling, and calibration leads to superior cross-domain generalization and early-stage detection capability.

Table 7. Ablation results of different loss function configurations

| Dataset | Model | MCI VS AD | | | AD vs NC | | | NC vs MCI | | |
|---|---|---|---|---|---|---|---|---|---|---|
| | | ACC | SEN | AUC | ACC | SEN | AUC | ACC | SEN | AUC |
| ADNI | $L_{KL}$ | 58.42 | 57.41 | 58.27 | 70.27 | 22.45 | 60.73 | 65.31 | 33.33 | 55.32 |
| | $L_{MSE}$ | 61.39 | 48.15 | 61.96 | 68.92 | 16.33 | 66.15 | 68.71 | 39.58 | 62.48 |
| | $L_{UCE}$ | 75.25 | 61.11 | 82.43 | 89.19 | 71.43 | 90.87 | 75.51 | 41.67 | 73.04 |
| | $L_{UCE}+L_{MSE}$ | 73.27 | 55.56 | 79.71 | **90.54** | 73.47 | **91.75** | 74.15 | 37.50 | 67.72 |
| | $L_{KL}+L_{MSE}$ | 59.41 | 59.26 | 60.64 | 75.00 | 26.53 | 72.79 | 69.39 | 8.33 | 56.36 |
| | $L_{KL}+L_{UCE}$ | 77.23 | 72.22 | 81.36 | 88.51 | 77.55 | 91.65 | 74.83 | 47.92 | **75.74** |
| | $L_{KL}+L_{UCE}+L_{MSE}$ | **80.20** | **83.33** | **83.57** | **90.54** | **83.67** | 91.34 | **77.55** | **56.25** | 73.46 |

## 5.3 Validation of Gaussian Distribution of Domain-Level Features

To verify the Gaussian distribution of domain-level style features used in the KL-divergence–based alignment loss, we analyzed the aggregated feature distributions from the ADNI, AIBL, and OASIS datasets. Figure 3, 4 and 5 shows the Q–Q plots (Quantile–Quantile Plot, a graphical method for comparing whether an empirical distribution follows a theoretical one; when the plotted points lie near the diagonal line, it indicates that the empirical distribution approximates the theoretical Gaussian distribution) of the first two principal components (PC1 and PC2, which represent the first and second principal components obtained through principal component analysis (PCA) and capture the dominant directions of variance in the style features within each domain) derived from the aggregated feature space. In each Q–Q plot, the horizontal axis represents the theoretical quantiles of a standard normal distribution, while the vertical axis represents the observed quantiles of the empirical feature distribution along each principal component. The data points closely follow the diagonal reference line, indicating that the empirical distributions of domain-level features exhibit near-Gaussian behavior. This observation empirically supports the Gaussian approximation assumed in Eq. (10) and justifies the use of the closed-form KL divergence between two Gaussian densities for measuring inter-domain distributional discrepancies. Consequently, this validation provides both theoretical and experimental evidence for aligning the mean and variance of domain-level style features in a consistent and statistically sound manner.

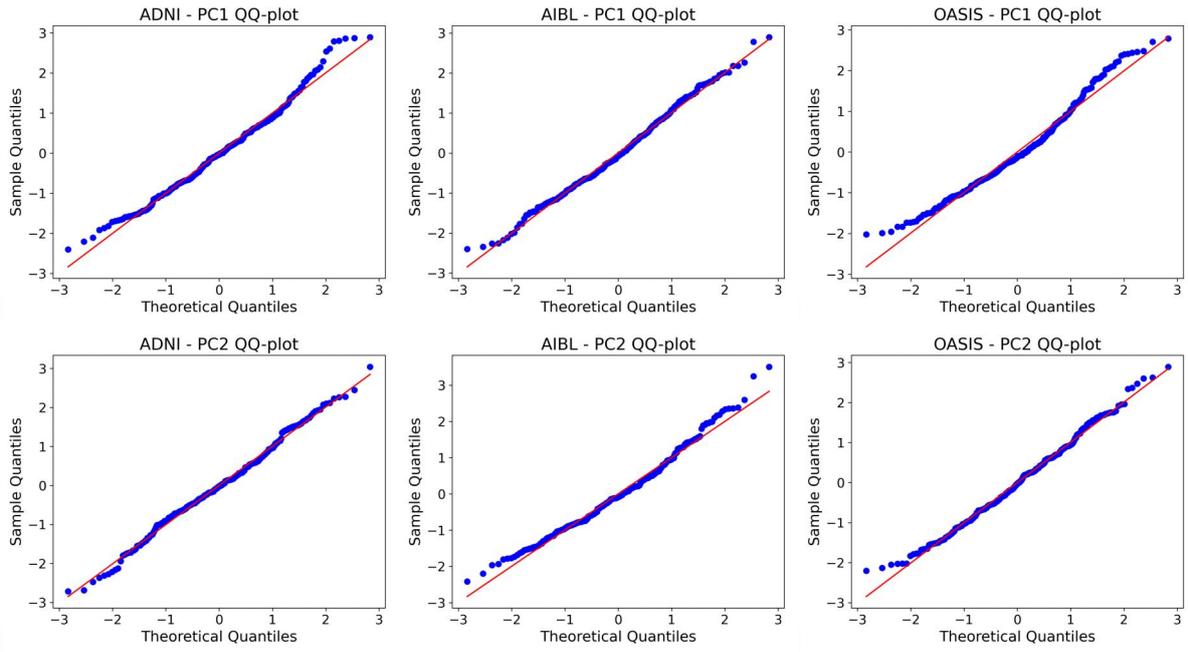

Figure 3. Q–Q plots of PC1 and PC2 for domain-level style features(NC VS. AD)

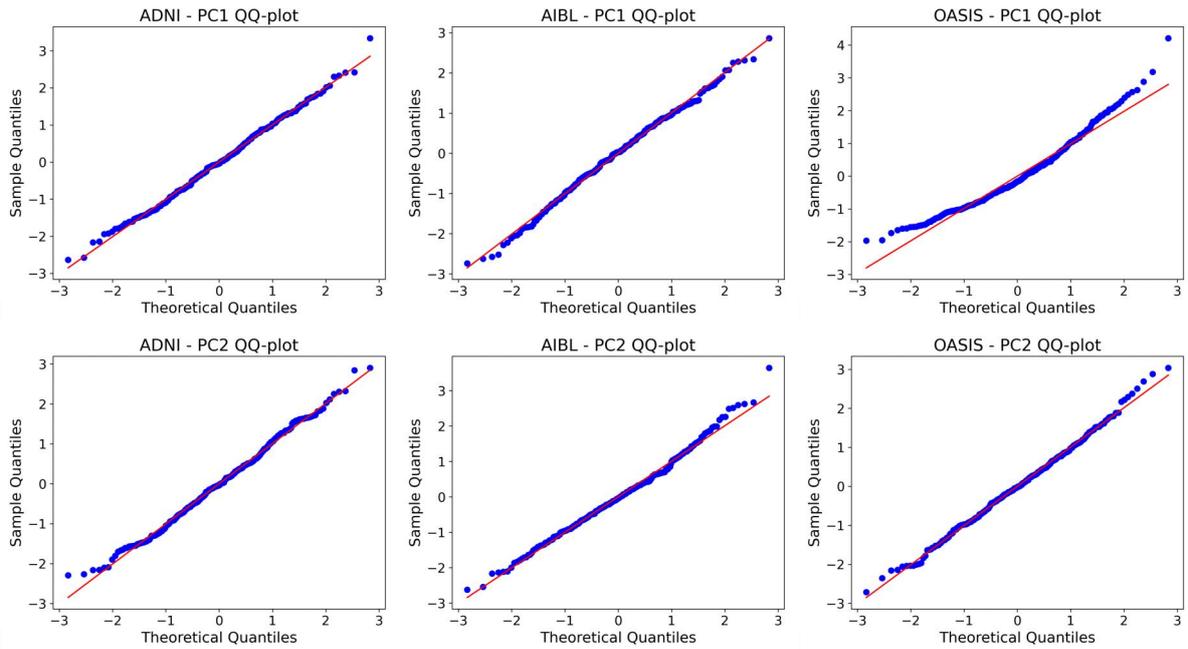

Figure 4. Q–Q plots of PC1 and PC2 for domain-level style features(NC VS. MCI)

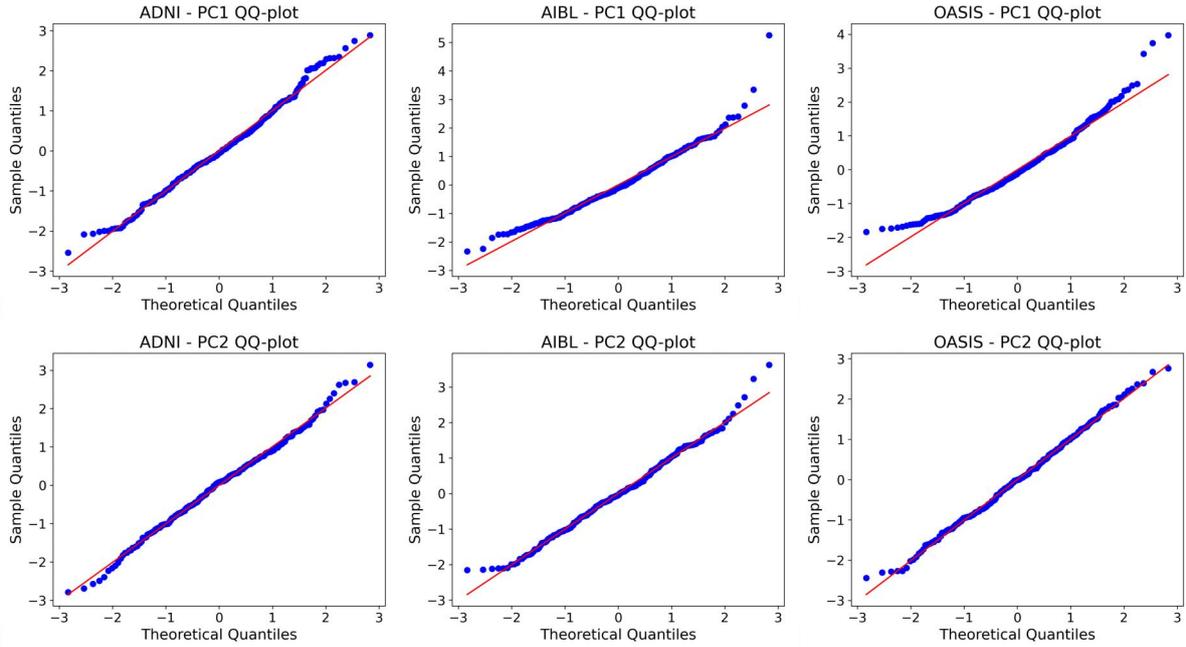

Figure 5. Q–Q plots of PC1 and PC2 for domain-level style features(MCI VS. AD)

### 5.4 Impact of Domain Adaptation on Cross-Center Classification Performance

Both models, with and without domain adaptation (DA), were trained under the same federated learning framework to ensure a fair and controlled comparison. In the No DA configuration, the uncertainty guided style alignment module was removed, while all other training and aggregation procedures remained identical. This setup isolates the specific contribution of the proposed DA strategy to cross center generalization. As shown in Table 8 and Figure 4, incorporating DA consistently improved performance across all binary classification tasks on the ADNI dataset. The comparison in Table 8 also highlights the substantial contribution of the DA mechanism to cross domain robustness within federated learning. Although the No DA model already achieved strong performance on the NC vs AD task, adding DA further increased sensitivity and provided more reliable identification of AD cases under inter center variability. The benefits of DA became even more evident in classification tasks involving earlier disease stages. In the NC vs MCI task, DA increased sensitivity from 39.58% to 56.25% and improved accuracy, showing that aligning domain specific representations enhances the detection of subtle cognitive decline that is often obscured by site differences. For the MCI vs AD task, DA produced a substantial gain in sensitivity, rising from 61.11% to 83.33%, together with an improvement in accuracy. These results indicate that domain adaptation reduces distributional mismatch across centers and stabilizes the learning of progression related biomarkers, leading to more reliable diagnostic performance across heterogeneous clinical environments.

Table 8. Results of ADNI dataset on different classification tasks (NO: not using domain adaptation; YES: using domain adaptation)

| ADNI | | ACC | SEN | AUC |
| --- | --- | --- | --- | --- |
| NC vs. AD | NO | **91.22** | 77.55 | 91.03 |
| | YES | 90.54 | **83.67** | **91.34** |
| NC vs. MCI | NO | 76.87 | 39.58 | **77.42** |
| | YES | **77.55** | **56.25** | 73.46 |
| MCI vs. AD | NO | 76.24 | 61.11 | **85.19** |
| | YES | **80.20** | **83.33** | 83.57 |

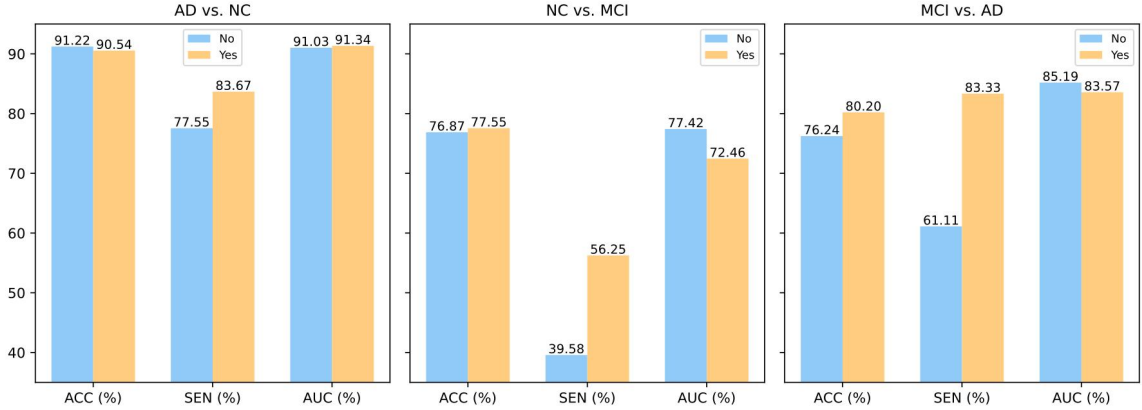

Figure 4. Impact of Domain Adaptation on ADNI Cross-Center Classification

### 5.5 Analysis of the Role of Uncertainty Quantification

The results in Table 10 highlight the value of uncertainty estimation as a mechanism for assessing reliability during inference. Low-uncertainty samples consistently produced the strongest performance, with accuracy peaks at the 0.2 threshold across all domains and tasks, including perfect AD vs NC accuracy on ADNI and over 90% accuracy on multiple MCI vs AD settings. The monotonic degradation in accuracy as uncertainty increased shows that the model assigns higher uncertainty to samples that are inherently more difficult or ambiguous. The stabilization observed at higher thresholds further suggests that uncertainty serves as a principled filter for identifying trustworthy predictions. These findings underscore the importance of integrating uncertainty estimation into domain-adaptive medical classification pipelines, enabling more robust and interpretable decision-making in heterogeneous clinical environments.

Table 10. Target Domain Classification Performance under Different Uncertainty Thresholds

| Target domain | MCI VS AD | | | AD VS NC | | | NC VS MCI | | |
|---|---|---|---|---|---|---|---|---|---|
| | ADNI | AIBL | OASIS | ADNI | AIBL | OASIS | ADNI | AIBL | OASIS |
| 0.2 | **90.48** | **90.00** | 93.33 | **100** | 88.57 | 90.00 | **88.24** | **89.19** | 87.50 |
| 0.4 | 87.50 | 81.25 | 83.78 | **100** | 89.22 | 91.23 | 86.11 | 81.67 | **88.29** |
| 0.6 | 88.24 | 83.33 | 80.00 | 96.88 | 88.39 | 88.89 | 83.33 | 79.49 | 85.49 |
| 0.8 | 86.11 | 79.31 | 79.43 | 93.65 | 86.07 | 85.71 | 84.34 | 73.28 | 84.87 |
| 1 | 80.20 | 73.03 | 79.72 | 90.54 | 86.30 | 79.01 | 78.23 | 72.67 | 83.73 |

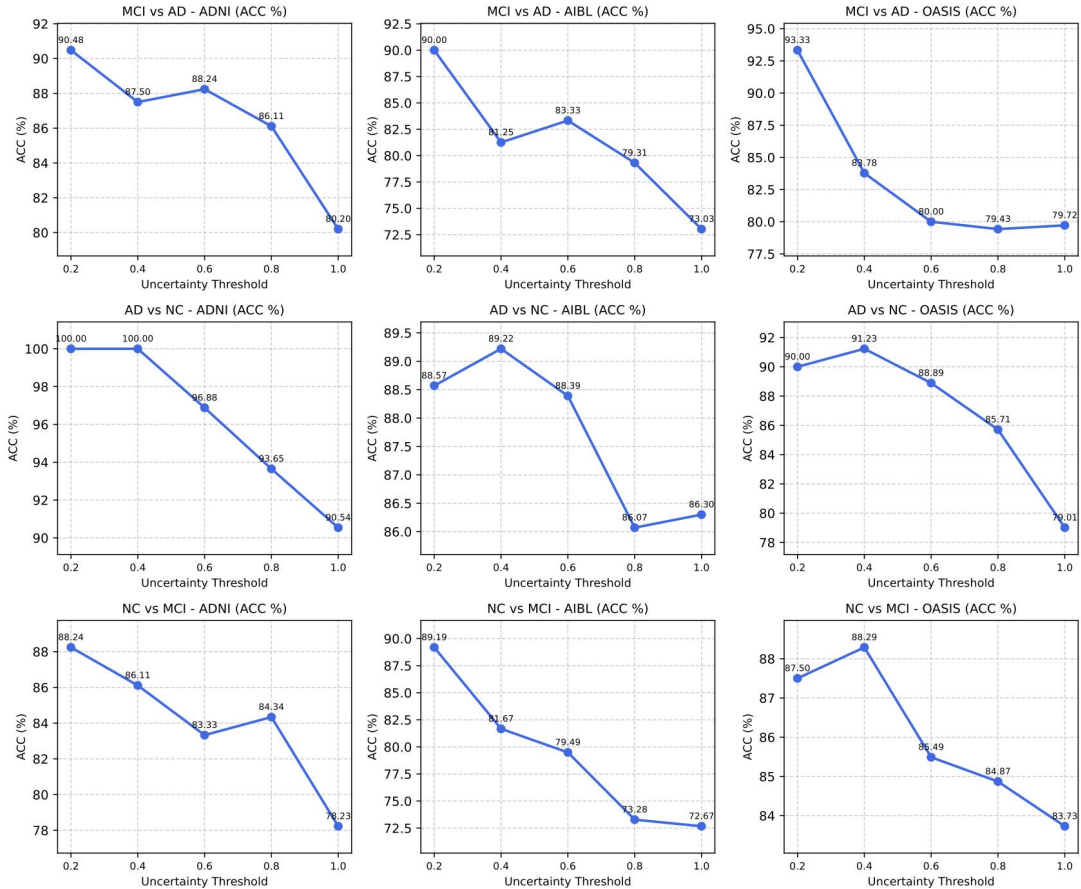

Figure 5. Target domain classification accuracy under different uncertainty thresholds.

### 5.6 Uncertainty-Weighted Aggregation

The comparison in Table 11 highlights the importance of integrating uncertainty-weighted aggregation into federated learning. Relative to the FedAvg baseline, the proposed weighting strategy substantially increased sensitivity across all tasks, most notably in NC vs AD (69.39% to 83.67%) and MCI vs AD (81.48% to 83.33%), reflecting a more reliable identification of clinically relevant cases. Improvements in accuracy across the three tasks further indicate that uncertainty weighting reduces the negative influence of clients with noisier or less informative distributions and promotes more stable global updates. Notably, the consistent gains observed in the MCI vs AD task suggest that uncertainty-aware aggregation is particularly effective when dealing with subtle disease progression signals. These findings demonstrate that incorporating uncertainty into aggregation criteria enhances robustness to inter-center heterogeneity and leads to more adaptive and trustworthy federated models.

Table 11. The impact of task federation uncertainty weighting on model generalization performance on the ADNI dataset (NO: FedAvg; YES: Our)

| ADNI | | ACC | SEN | AUC |
|---|---|---|---|---|
| NC vs. AD | NO | 89.19 | 69.39 | **92.25** |
| | YES | **90.54** | **83.67** | 91.34 |
| NC vs. MCI | NO | 74.83 | 54.17 | **76.20** |
| | YES | **77.55** | **56.25** | 73.46 |
| MCI vs. AD | NO | 74.26 | 81.48 | 75.37 |
| | YES | **80.20** | **83.33** | **83.57** |

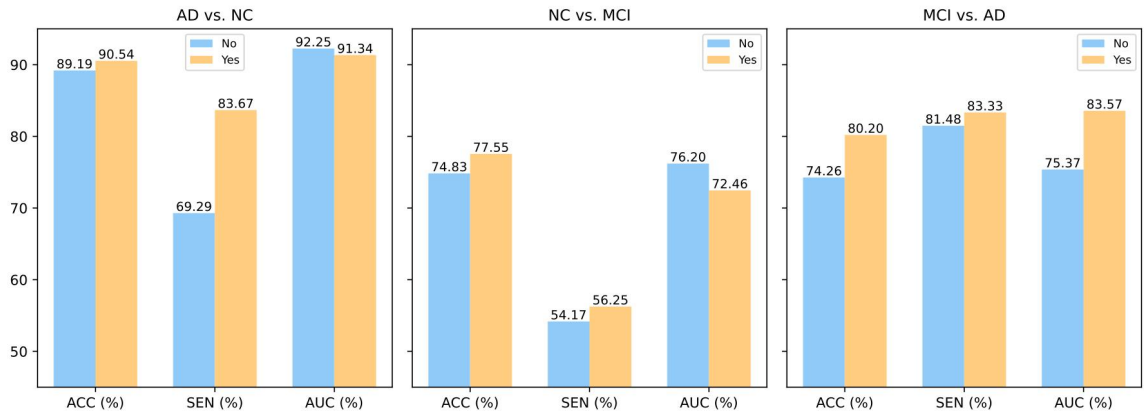

Figure 6. Impact of uncertainty-weighted task federation on model generalization across ADNI classification tasks.

### 5.7 Feature Importance Analysis for NC, MCI, and AD Classification

(1) Feature Importance Analysis for NC vs. MCI Classification

In the NC vs. MCI feature-importance analysis, we applied integrated gradients (IG) and occlusion (OCC), normalized their scores (IG_norm, OCC_norm), averaged them into a joint score, and labeled RoIs with joint score > 0.5 as highly discriminative.The top-ranked feature was the right superior temporal gyrus (joint score = 0.934), which supports higher-order auditory and language functions and has been linked to early temporal-lobe changes in MCI, plausibly associated with β-amyloid deposition and tau pathology. The second-ranked feature was the corpus callosum (joint score = 0.815), reflecting early alterations in white-matter connectivity in MCI and aligning with diffusion tensor imaging (DTI) evidence of reduced callosal integrity. The right pulvinar of the thalamus (joint score = 0.710) participates in sensory–cognitive integration, is implicated in thalamocortical disconnection in MCI, and accords with resting-state fMRI evidence of default mode network (DMN) disruption. Other important features included the left basolateral amygdala (joint score = 0.657), implicated in emotion and memory; the right anterior cingulate gyrus (0.607), a DMN hub; the right dorsal insula (0.3–0.5), involved in emotion and interoception; and the cerebellar vermis (< 0.3), which showed lower discriminative utility. These findings indicate that NC–MCI classification relies primarily on temporal-lobe, callosal, thalamic, and DMN-related regions, reflecting early functional and structural alterations in MCI.As shown in Figure 7, a brain heatmap for NC vs. MCI visualizes the spatial distribution of feature importance.

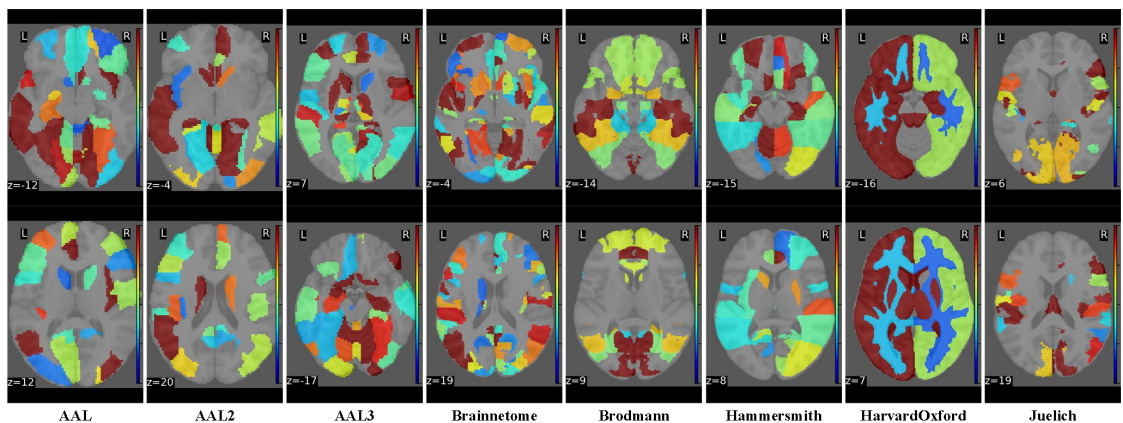

Figure 7. Feature Visualization Heatmap of NC vs. MCI

(2) Feature Importance Analysis for MCI vs. AD Classification

In the MCI vs. AD task, high-importance features (joint score > 0.5) highlighted regions implicated in progressive neurodegeneration in AD. The top-ranked feature was the corpus callosum (joint score = 0.892), reflecting further deterioration of white matter connectivity in AD and aligning with. The mid-portion of the left temporal pole (joint score = 0.876) supports semantic memory and social cognition and has been associated with temporal-lobe atrophy in AD, consistent with voxel-based morphometry (VBM) studies. The right posterior superior temporal sulcus (joint score = 0.789) supports social cognition and language processing and has been linked to temporo-parietal dysfunction in AD. Right Broca's area (BA45; joint score = 0.645) was implicated in language production and aligned with the language decline observed in AD. The middle temporal gyrus, temporo-occipital part (joint score = 0.612), reflects visual and semantic processing impairments; as a core default mode network (DMN) region, its functional connectivity has been reported to be markedly reduced in AD.The left inferior temporal gyrus (joint score = 0.3–0.5) was implicated in memory and language processing, indicating more widespread cortical involvement. Lower-ranked features showed limited specificity for distinguishing MCI from AD.As shown in Figure 8, a brain heatmap visualizes the spatial distribution of feature importance for MCI vs. AD.

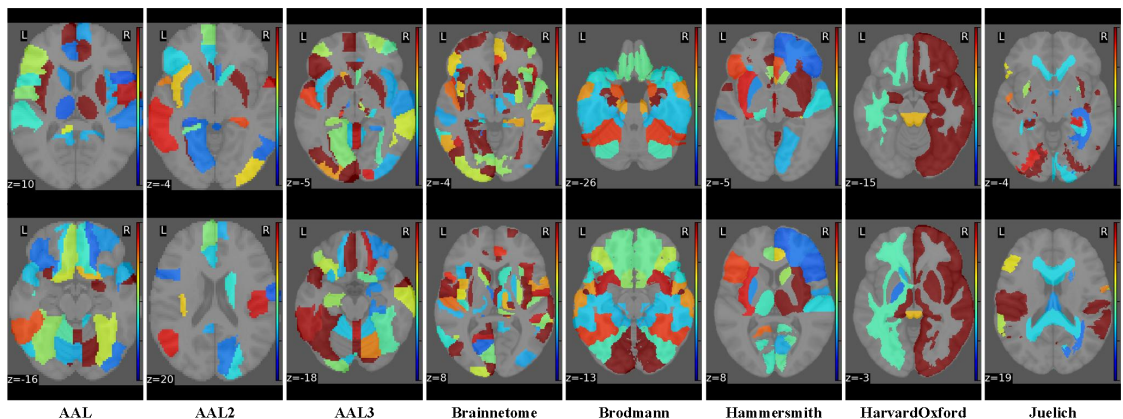

Figure 8. Feature Visualization Heatmap of MCI vs. AD

(3) Feature Importance Analysis for NC vs. AD Classification

In the NC vs. AD task, feature-importance analysis revealed RoIs indicative of widespread late-stage neurodegeneration in AD. The corpus callosum (joint score = 0.925) ranked first, indicating severe white matter connectivity damage in AD and aligning with structural MRI and diffusion tensor imaging (DTI) evidence.The mid-portion of the left temporal pole (joint score = 0.904) supports semantic memory and social cognition and has been associated with widespread temporal-lobe atrophy in AD. The right posterior superior temporal sulcus (joint score = 0.867) highlighted deficits in social cognition and language. Right Broca's area (joint score = 0.733) was implicated in language production, consistent with language impairment in AD. The right superior temporal gyrus (joint score = 0.689) supports higher-order auditory and language processing. These findings are consistent with reports from ADNI of cortical thinning and functional decline in late-stage AD. Mid-ranked features (joint score = 0.3–0.5) involved default mode network (DMN) and emotion-processing regions. Lower-ranked features showed weaker associations with AD-specific pathology. As shown in Figure 9, a brain heatmap for NC vs. AD visualizes the spatial distribution of feature importance.

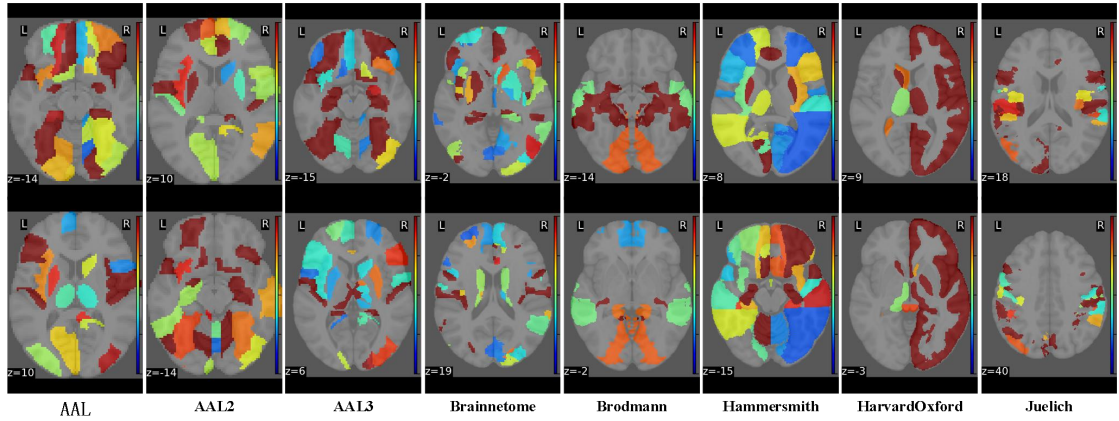

Figure 10. Feature Visualization Heatmap of NC vs. AD

Across the three pairwise tasks—NC vs. MCI, MCI vs. AD, and NC vs. AD—the corpus callosum (joint scores = 0.815, 0.892, and 0.925, respectively) and temporal-lobe regions ranked highly, underscoring their central role along the NC–MCI–AD continuum and spanning white-matter connectivity, semantic memory, and language processing. Temporo-parietal regions (e.g., the right posterior superior temporal sulcus) and default mode network (DMN)–related areas (e.g., the right anterior cingulate gyrus and the middle temporal gyrus, temporo-occipital part) were especially prominent for MCI vs. AD and NC vs. AD, highlighting their importance in later disease stages. For NC vs. MCI, the right thalamic pulvinar and the left basolateral amygdala were emphasized, reflecting early deficits in sensory–cognitive integration and emotion processing. Across all tasks, highly weighted features (joint score > 0.5) clustered in the corpus callosum, temporal lobe, temporo-parietal regions, and DMN-related structures, consistent with reports from ADNI of progressive white-matter damage, cortical atrophy, and functional decline in AD. These findings indicate that the model captures biologically plausible signals and suggest that top-ranked RoIs may serve as candidate biomarkers for early screening (NC vs. MCI), staging (MCI vs. AD), and late-stage diagnosis (NC vs. AD).

## 6. Conclusion

We present a classification framework for Alzheimer's disease that integrates uncertainty quantification with federated domain adaptation to enhance target-domain learning and improve cross-domain generalization. A Kullback–Leibler divergence, weighted by predictive uncertainty, is minimized to mitigate distribution shifts between domains. In target-domain training, extracted target features are concatenated with source and target style features to stabilize the learning process. Federated aggregation further incorporates an uncertainty-aware weighting scheme that modulates each site's contribution to the global model. Experiments on multiple cross-domain tasks have demonstrated consistent improvements in target-domain classification, validating the efficacy of the proposed method.

**Acknowledgments**

This study received support from the National Natural Science Foundation of China (Grant Numbers: 62476255, 62303427, and 82370513), the Science and Technology Innovation Talent Project of Henan Province University (Grant Number: 25HASTIT028), the Henan Province Key Scientific Research Projects Plan for Higher Education Institutions (Grant Number:26A520044), and Zhongyuan Science and the Technology Innovation Outstanding Young Talents Program